%% file: lodos.tex
\pdfoutput=1
\documentclass[10 pt, conference,capitalise]{ieeeconf}  %

\IEEEoverridecommandlockouts                %

\input{header.tex}

\usepackage[T1]{fontenc}

\usepackage{cinzel}
\DeclareRobustCommand{\Lodos}{%
  {\color{ourdarkblue2} \cinzel \selectfont LODOS}%
}

\definecolor{plotgray}{HTML}{4A4A4A}

\newcommand{\plotlabel}[3]{%
  \node[
    anchor=north west,
    inner sep=0pt,
    text=plotgray,
    font=\sffamily\bfseries\fontsize{6}{8}\selectfont %
  ] at ([xshift=#1,yshift=-#2]plot.north west) {#3};
}

\title{\LARGE \bf \Lodos: A Lightweight Simulator Designed for \\ Contact-Rich Interactions with Detailed Meshes}

\author{\normalsize
        Onur Beker$^{1}$,
        A.\ Ren\'e Geist$^{1}$,
        Georg Martius$^{1,2}$ \\
        ${}^{1}$University of Tübingen, ${}^2$Max Planck Institute for Intelligent Systems
        }
\begin{document}

\IEEEaftertitletext{%
    \vspace{-3mm}
    \centering
    \begin{tikzpicture}
        \node[inner sep=0pt, outer sep=0pt] (plot) {%
            \includegraphics[width=\linewidth]{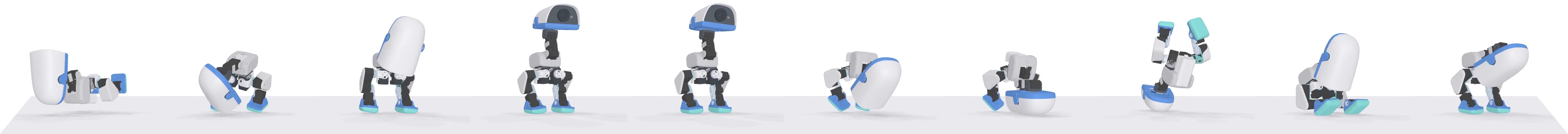}%
        };
        \plotlabel{4mm}{0mm}{MJWarp}
    \end{tikzpicture} \\
    \vspace{4mm}
    \begin{tikzpicture}
        \node[inner sep=0pt, outer sep=0pt] (plot) {%
            \includegraphics[width=\linewidth]{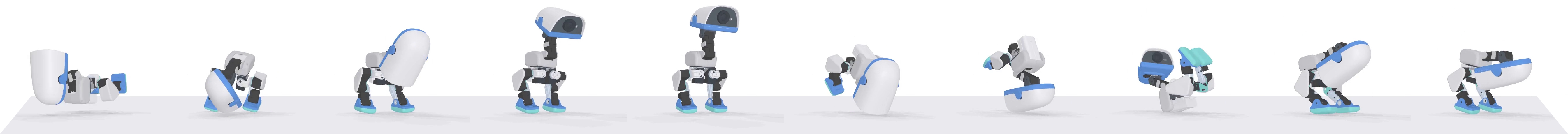}%
        };
        \plotlabel{4mm}{0mm}{Lodos-MJ-2}
    \end{tikzpicture}\\
    \vspace{0.1mm}
    \captionof{figure}{A \href{https://github.com/pollen-robotics/microduck}{Microduck} getting up and rolling, simulated in MJWarp and Lodos. The policy is trained in MJWarp and transferred zero-shot.}
    \label{fig:overview}
    \vspace{0.5\baselineskip}
}

\maketitle

\begin{abstract}
In recent years, massively parallelized rigid-body simulation has played a central role in robotics research, enabling rapid progress in areas such as reinforcement learning, stochastic optimal control, and randomized smoothing-based trajectory optimization. Particularly in contact-rich settings, the scalability of such simulations in large part depends on well-optimized representations of collision geometry, often requiring a coarse approximation of the underlying surfaces using primitives such as spheres, capsules, cylinders, and boxes. Whenever the complexity of the surfaces or the precision demands of the application render such approximations insufficient, it is instead necessary to employ mesh-mesh collisions, which impose a significant reduction in throughput. This paper proposes a simulation framework designed from scratch to address this bottleneck, with novel routines for broadphase and narrowphase collision detection as well as the subsequent contact dynamics computation. As experimentally validated, the framework matches existing well-established simulators in terms of physical faithfulness, and can achieve approximately an order of magnitude higher throughput in the detailed mesh-mesh collision setting that it specializes in. The source code is available at \href{https://github.com/martius-lab/lodos}{https://github.com/martius-lab/lodos}.
\end{abstract}

\section{Introduction} \label{intro}
What makes a good simulator for solving control problems that involve contact-rich interactions? Let us build the desiderata by focusing on the control and contact-rich interaction parts of the question separately.

{\color{ourblue}{\textbf{A good simulator for control:}}} Sampling-based control approaches that leverage massively vectorized simulation \cite{makoviychuk2021isaac, MuJoCo_warp} have proven to be very proficient at solving contact-rich manipulation tasks. A common pattern shared by such approaches, whether it is domain randomization with RL \cite{serifi2024vmp, allshire2022transferring}, risk-aware control with stochastic MPC \cite{kurtz2024hydrax}, or randomized smoothing \cite{pang2023global} with gradient-based optimization, is that actions are selected such that they remain \say{approximately correct} with respect to metrics averaged over randomized ensembles of contact parameters generated via artificial (and oftentimes non-physical) perturbations (e.g. adding virtual forces, noisy friction coefficients and haptic sensing, approximate or domain randomized collision geometry). And a main argument of this paper is that in such a setting, perturbations and averaging already introduce a \say{convex-hull} effect on how much the physical faithfulness of each ensemble member can reflect on (or matter for) the selection of the common actions facing the ensemble. Therefore, there should be room for using simplified versions of contact dynamics (that can still sufficiently preserve the \say{convex-hull}) to achieve drastically increased ensemble sizes instead, resulting in a net benefit for the downstream control algorithm.

{\color{ourblue}{\textbf{A good simulator for contact-rich interactions:}}} Most of the commonly used simulators in robotics research \cite{todorov2012MuJoCo, drake, carpentier2024compliant, tsounis2026kamino} use a combination of convex polytope based collision detection \cite{gilbert1988fast, gregorius2015robust} with a numerical optimization routine to solve the regularized non-linear complementarity (NCP) or cone-complementarity (CCP) systems of equations describing contact dynamics \cite{andrews2022contact, castro_sim, lelidec2023contacts, carpentier2024compliant, anitescu2006optimization, stewart2000rigid}. While this battle-tested recipe certainly has its merits, it suffers from the following drawbacks particularly in contact-rich settings:
\begin{itemize}
    \item The regularization introduced to the NCP/CCP equations scales frictional contact forces down from the exact value needed to maintain complete staticity in a rigid-body system (e.g. zero velocity and acceleration), meaning all contacts eventually slip under constant forces such as gravity.
    \item Applying GJK-EPA algorithms \cite{gilbert1988fast, van2003collision} to convex polytope collisions produces a single contact point, which is occasionally further extended into a sparse set of contact points lying at the corners of a 2D polygon \cite{gregorius2015robust}. Neither mechanism can faithfully implement contact forces distributed over a patch, meaning torsional friction is emulated in an ad hoc manner by introducing artificial constraints to the optimization routine that oppose motion around the contact normal.
    \item Convex-polytope-based collision routines scale poorly with parallelization, particularly under contact-rich configurations involving complex non-convex meshes.
\end{itemize}
In concert, these drawbacks chart a path toward the design of an effective simulator for contact-rich interactions.

{\color{ourblue}{\textbf{Contributions.}}} We present a novel simulation routine designed from scratch to facilitate research on developing control algorithms in contact-rich settings. In particular:
\begin{itemize}
    \item We propose a new collision detection routine built upon sphere - SDF collisions as the axiomatic interaction, which can represent arbitrary non-convex (closed) surfaces at arbitrary levels of precision, and scales much more favorably with massive parallelization compared to the traditional convex polytope based approach. Specifically, this includes a novel hierarchical broadphase routine to logarithmically cull potential sphere-SDF interactions, and a novel narrowphase routine using 3D-spline-based SDF and surface normal representations whose evaluation time is independent of geometry complexity.
    \item We propose a new model of contact dynamics based on rescaling spring-damper forces via a staggered fixed-point iteration scheme, derived from a lumped parameterization of the traditional SOCP-based contact models employed in most common simulators \cite{anitescu2006optimization, todorov2012MuJoCo}. The main benefits are that it: i) scales much more favorably under massive vectorization compared to standard approaches, ii) does not exhibit a slipping artifact, iii) allows patch-patch contacts and therefore faithful torsional friction simulation.
\end{itemize}

\begin{figure}
    \centering
    \begin{minipage}[c]{0.60\linewidth}
        \centering
        \includegraphics[width=\linewidth]{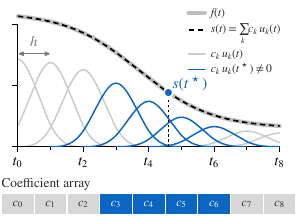}
    \end{minipage}\hfill
    \begin{minipage}[c]{0.4\linewidth}
        \centering
        \includegraphics[width=0.7\linewidth]{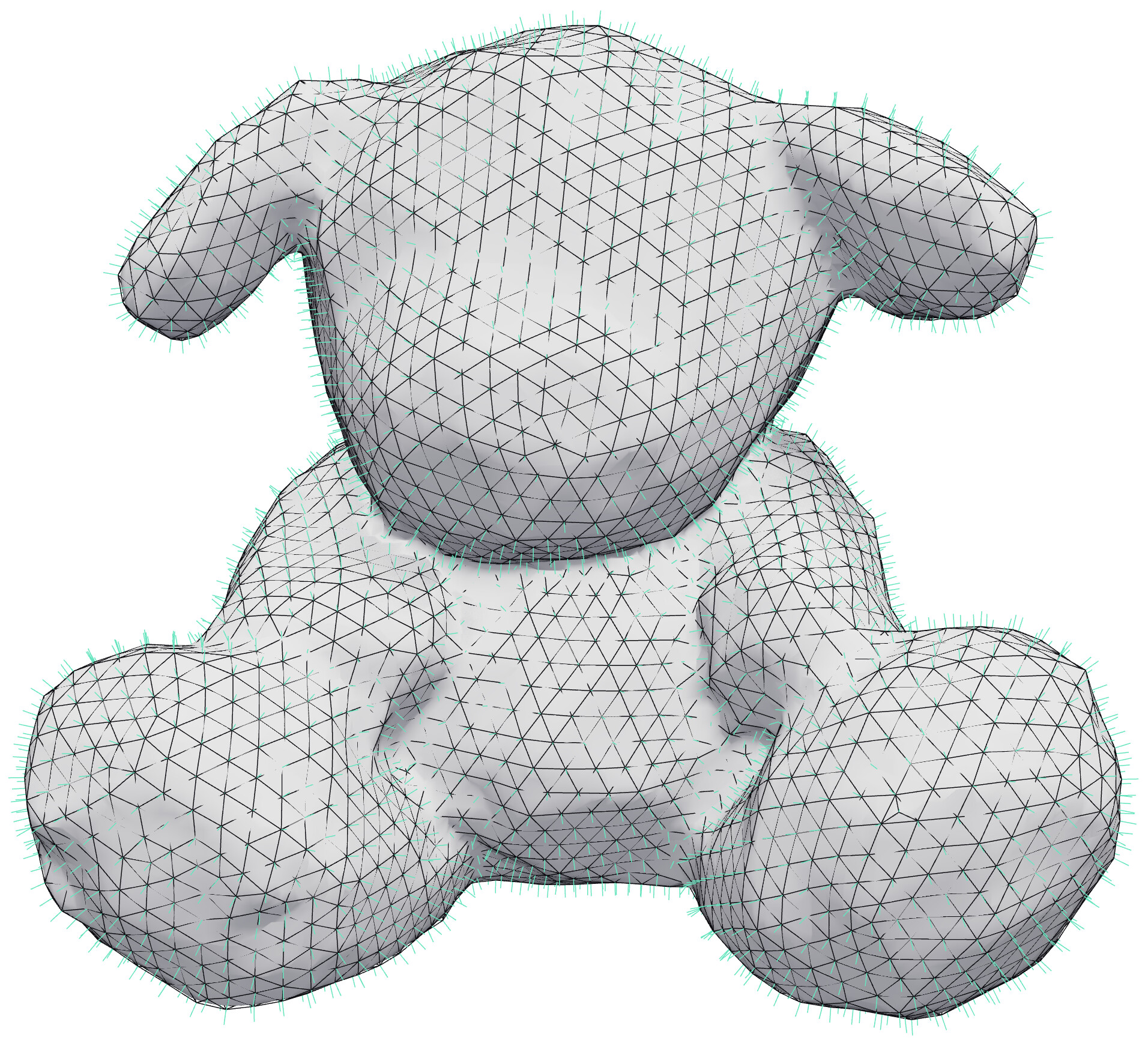}\\
        \includegraphics[width=0.6\linewidth]{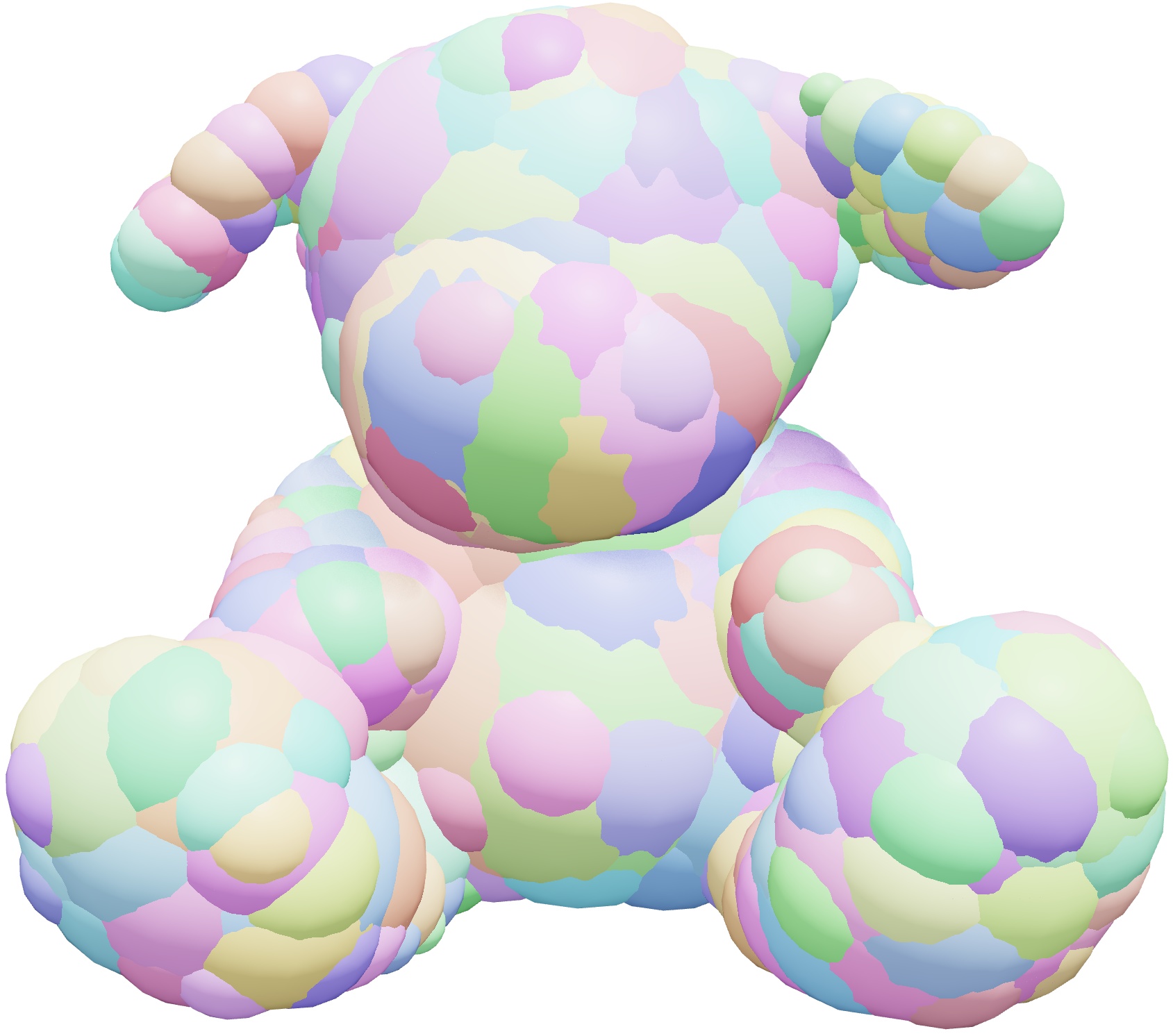}
    \end{minipage}
    \vspace{-1mm}
    \caption{\textbf{Left:} Cubic spline interpolation. In 1D, evaluating the spline $s(t^{\star})$ requires accessing only four entries in the coefficient array (more generally $4^N$ in $N$-D). \textbf{Right:} A collision geometry is represented via two pieces of information: i) an SDF implemented with a 3D spline, and ii) zero volume spheres (i.e. vertices) densely distributed on the 2D surface of the mesh and inscribed spheres distributed within the 3D volume of the mesh.
     }
    \label{fig:spline-illustration}
    \vspace{-4mm}
\end{figure}

\vspace{-2mm}
{\section{Related Work}} \label{related_work}
\vspace{-1mm}
Rigid-body simulation with contact is central to robotics. Classical methods, still used in CAD software \cite{mscadams}, model contact with dissipative spring-dampers \cite{andrews2022contact, hunt1975coefficient, kurtz2025cenic}, yielding stiff ODEs that require slow, time-adaptive integration. Discrete time-stepping methods \cite{stewart1996implicit, stewart2000rigid} instead jointly determine contact forces and post-impact velocities through the NCP equations \cite{lelidec2023contacts}. Appropriate relaxations yield efficient convex approximations \cite{anitescu2006optimization}, as used in MuJoCo \cite{todorov2012MuJoCo}. Alternatively, fixed-point iterations with a \say{Signorini correction} can solve the original NCP system \cite{carpentier2024compliant}, as in Simple \cite{simple_robotics_simple}.
A recent, complementary direction is the implementation of such models of contact dynamics on the GPU, to create massively parallelized simulators such as IsaacGym \cite{makoviychuk2021isaac}, MJWarp \cite{MuJoCo_warp}, and Kamino \cite{tsounis2026kamino}. This paper shares the same goal as such frameworks: to maximize simulator throughput. However, unlike these frameworks that aim to provide high-quality GPU implementations while remaining completely faithful to existing, well-established models of contact dynamics (that were not necessarily formulated with massive parallelization in mind), the proposed approach sets out to create a new contact model that was designed from scratch with massive parallelization as the main concern, specifically in settings involving collisions between detailed non-convex meshes.

\section{Methods: Collision Detection} \label{methods_col}
Sphere-SDF collisions are the axiomatic interaction around which the entire collision detection stack is built. Therefore, we start with a description of how we represent SDFs.
\subsection{Efficiently Representing SDFs with 3D Splines}
\subsubsection{Cubic Splines in 3D}
In 1D, standard cubic splines are built as a sum of twice-differentiable polynomial basis functions $u: \mathbb{R} \to \mathbb{R}$ of the following form \cite{habermann2007multidimensional}:
\vspace{-2mm}
\begin{align}
    u(t) = {\footnotesize
    \begin{cases}
    4 - 6|t|^2 + 3|t|^3 \quad &\text{for} \ |t| \leq 1  \\
    (2 - |t|)^3  \quad &\text{for} \ \ 1 < |t| \leq 2 \  \\
    0 \quad &\text{for} \ 2 < |t|
    \end{cases}}
\end{align}
\vspace{-0.5mm}
Given the values $\{f_i\}_{i=0}^N$ of a function $f: [a, b] \to \mathbb{R}$ evaluated at the corners of $N$ intervals splitting $[a, b]$ into equal pieces of length $h$ each, the cubic spline interpolation in 1D is given by the sum $s(t) = \sum_{k=1}^{N + 3} c_k u_k(t)$, using a superposition of $N+3$ shifted basis functions defined as $u_i(t) = u(\frac{t - a}{h} - k + 2)$. The coefficients $c_k$ are in turn obtained by solving the linear system defined by the equations $s(t_i) = f_i$.
Given the values $\{f_{i, j, k}\}_{i, j, k=0}^{N_x, N_y, N_z}$ of a function $f: \mathbb{R}^3 \to \mathbb{R}$ evaluated at the corners of a 3D voxel grid with $(N_x, N_y, N_z)$ voxels along each axis, a 3D spline interpolation is obtained as $s(x, y, z) = \sum_{k_x=0}^{N_x+3}\sum_{k_y=0}^{N_y+3}\sum_{k_z=0}^{N_z+3} \ c_{k_x, k_y, k_z} \ u_{k_x}(x) u_{k_y}(y) u_{k_z}(z)$, using products of the form $u_{k_x}(x) u_{k_y}(y) u_{k_z}(z)$ to define 3D basis functions from 1D basis functions. The coefficients $c_{k_x, k_y, k_z}$ are again obtained by solving the linear system defined by the equations $s(x_i, y_j, z_k) = f_{i, j, k}$.
The compact support of $u(t)$ ($u(t)=0$ outside $[-2,2]$) makes evaluation fast and independent of grid size. In 3D, only $64$ terms of $s(x,y,z)$ are nonzero at any query point, and their indices can be identified via integer rounding with minimal computational cost.

\subsubsection{Using 3D Splines to Represent SDFs and Normals} We interpolate mesh signed distances sampled at voxel corners with a 3D cubic spline. Its normalized gradient (computed in closed form) defines the surface normal function (SNF). Compact basis support makes SDF and SNF evaluation costs independent of geometry complexity, which affects only the required grid resolution and storage. The grid is shared across parallel instances through concurrent GPU reads, so its storage does not grow with batch size.

\subsection{Narrowphase Collision Detection}
The goal of the narrowphase is to generate a set of contact points together with penetration depths and contact normals. Let us start with a description of the axiomatic sphere-SDF collision, around which the rest of the routine is built.

\subsubsection{Sphere-SDF Collisions} Let $\bm c \in \mathbb R^3$ be the center of a sphere with radius $r \in \mathbb R$, colliding with a surface with the SDF $\phi(\bm x): \mathbb R^3 \to \mathbb R$ and SNF $\bm n(\bm x): \mathbb R^3 \to \mathbb R^3$. The contact point (lying on the surface of the sphere) is $\bm p = \bm c - \bm n(\bm c) r$, with the contact depth $d = \phi(\bm c) - r$ (negative for penetration) and contact normal $\bm n = \bm n(\bm c)$.
An important observation is that a vertex--SDF collision \cite{beker2026smoothly} is strictly a special case of a sphere-SDF collision with $r = 0$.

\subsubsection{Narrowphase} As shown in Fig.\ref{fig:spline-illustration}, in the proposed framework, every collision geometry is represented by three pieces of information: a set of $K$ spheres $\mathcal{S} = \{(\bm c_k, r_k)\}_{k=1}^K$ that are distributed across its surface and inscribed within its volume, together with an SDF $\phi(\bm x)$ and an SNF $\bm n(\bm x)$. Collision pairs are asymmetric: when a geometry $i$ is colliding with a geometry $j$ (forming a directional collision pair $(i, j)$), it means the spheres $\mathcal{S}_i$ are colliding with the SDF and SNF $\phi_j, \bm n_j$ to produce $|\mathcal{S}_i|$ contact points $\bm p_{i, j, k}$ with depths $d_{i, j, k}$ and normals $\bm n_{i, j, k}$. Symmetry, if desired, is achieved by defining two collision pairs $(i, j)$ and $(j, i)$ in both directions.

Naturally, the fidelity of the scheme depends on how well the spheres are distributed. The key insight is the following. Exact mesh-SDF collisions can be implemented by considering all vertex--SDF and edge--SDF contacts \cite{beker2026smoothly}, and vertex--SDF contacts are already covered as a special case of sphere-SDF contacts. Therefore, we split the sphere placement $\mathcal{S}$ into two groups: i) zero-volume spheres (i.e. vertices, $r=0$) densely distributed on the 2D surface of the mesh (i.e. implementing vertex--SDF contacts), and ii) inscribed spheres distributed within the 3D volume of the mesh to cover as much volume as possible. To create the former group, we employ the pipeline from \cite{beker2026smoothly} to obtain a dense isotropic remesh and take the resulting vertices. The latter group can be constructed manually \cite{vanderheijden2024bubblify} (the option we opt for), or any sphere decomposition algorithm can be used \cite{2025foamtoolsphericalapproximation}. It is also important to note that there is no distinction between how the two different groups of spheres are processed by the software implementation of the broadphase and narrowphase routines (the former group simply has all radii set to $r=0$).

\textit{3) Why omit edge--SDF contacts?}
In the proposed framework, vertex--SDF and inscribed-sphere--SDF collisions are already exact, but omitting edge--SDF contacts can in theory admit millimeter-scale penetrations into the gaps between surface vertices and underlying spheres. This artifact can be detected and continuously measured by evaluating inscribed-sphere contact points against their own geometry's SDF (i.e. contacts should not lie far beneath that surface). It arises only: i) with pathological vertex distributions (e.g., coarse non-isotropic meshes), and ii) in edge--edge interactions, such as two cube edges meeting. Such edge-edge interaction modalities are inherently unstable and therefore undesirable for manipulation. And a controller can avoid them by penalizing the penetration measure. From a {\color{ourblue}\textit{physics-for-control perspective}}, this is preferable to a $\sim 2$--$4\times$ slowdown for edge--SDF detection that suppresses a minor, detectable, and avoidable artifact associated with undesirable behavior.

\subsection{Broadphase Collision Detection}\label{sec:broadphase}
Without any broadphase, narrowphase involves the spheres $\mathcal{S}_i$ of a geometry $i$ colliding with the SDF and SNF $\phi_j, \bm n_j$ of another geometry $j$ to produce $|\mathcal{S}_i|$ contact points. And while the speed and computational cost of $\phi_j, \bm n_j$ do not depend on the geometry complexity, the number of spheres $|\mathcal{S}_i|$ does and it can be as large as $\sim 10^4$ for very complex surfaces. The key observation that allows broadphase is that by definition, contact interactions keep the intersection between colliding surfaces minimal, meaning the overwhelming majority of the $|\mathcal{S}_i|$ contact points have positive signed distances and hence do not affect the underlying physics. These non-penetrating contact points can easily be detected and culled in an $\mathcal{O}(\log |\mathcal{S}_i|)$ manner using a hierarchical culling scheme as described in the upcoming sections.
\subsubsection{Voxelized SDF}
To enable the broadphase, in addition to the voxel-corner samples used for spline interpolation, we store signed distances at voxel centers for $\mathcal O(1)$ lookup. For a point $\bm p$ at distance $\Delta d$ from its voxel center $\bm p_c$, $\phi(\bm p)+\Delta d\geq\phi(\bm p_c)$ holds \cite{beker2026novel}. If $r_v$ is the voxel circumsphere radius, then $\phi(\bm p_c)-r_v$ bounds $\phi(\bm p)$ from below. Thus, a sphere of radius $r$ centered at $\bm p$ can be culled whenever $\phi(\bm p_c)-r_v\geq r$.

\begin{figure}
    \centering
    \begin{minipage}[c]{0.48\linewidth}
        \centering
        \includegraphics[width=\linewidth]{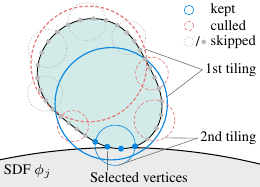}
    \end{minipage}\hfill
    \begin{minipage}[c]{0.48\linewidth}
        \centering
        \includegraphics[width=\linewidth]{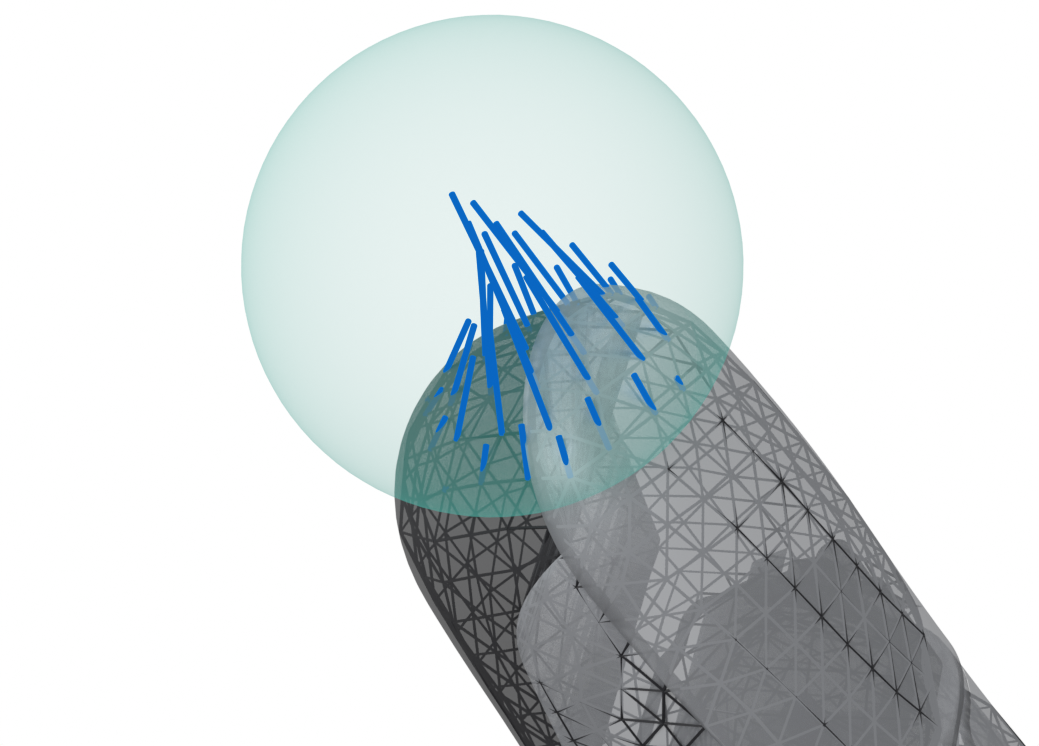}
    \end{minipage}
    \caption{\textbf{Left:} Selecting vertices for the narrowphase using tiled sphere hierarchies. \textbf{Right:} The resulting dense contact force field.
     }
    \label{fig:tiling-and-detection}
    \vspace{-5mm}
\end{figure}

\subsubsection{Tiled Sphere Hierarchies} To enable efficient culling of spheres $\mathcal{S}_i$ using the voxelized SDF checks from the previous section, we need to build a hierarchical acceleration structure. This is achieved as follows.
First, we serialize sphere centers along a 3D Hilbert space-filling curve, a standard locality-preserving technique in geometry processing \cite{morton1966computer, wu2024ptv3}.
This ensures that if the centers of two spheres are close in terms of 3D Euclidean distance, then their indices are also close together after reordering.
Then, $T$ consecutive spheres are grouped into a single \say{tile} (resulting in $\sim |\mathcal{S}_i| / T$ tiles), and the smallest bounding sphere of every tile is computed and stored. The number $T$ is fixed for every tile to make reindexing and stream compaction more efficient.
In principle, this process of serialization and tiling can be repeated an arbitrary number of times to construct a tile hierarchy.
In practice, we found two levels to be sufficient for geometries with several thousand spheres.
The broadphase traverses the hierarchy from the highest level. At each level, voxelized SDF checks cull tiles, and stream compaction records surviving indices for the next level. Since the dimensions of stream compacted arrays depend on geometry poses, all variable-size scans use a \say{reduce-then-scan} strategy \cite{ragankelley2025lecture7}. Surviving leaf spheres then enter the narrowphase.

\subsubsection{Preventing Host-Device Synchronization} While the reduce-and-scan strategy allows variable-length scans to run entirely within the GPU without any CPU communication, the memory that stores the arrays being scanned still needs to be preallocated at compile time to prevent host-synchronization and allow CUDA graph capture \cite{nvidia_cuda_guide}. And since an overwhelmingly small percentage of all sphere-SDF contacts is active during a simulation, allocating memory for the worst-case scenario of $|\mathcal{S}_i|$ contacts per simulation instance would be drastically wasteful. Therefore, the \say{capacities} that determine how much memory needs to be allocated (as a percentage of the estimates for all $|\mathcal{S}_i|$ contacts) are calibrated live as the simulation is running, via a simple control rule that monitors the largest number of contacts $C$ observed so far in the lifetime of a simulation. At the beginning of the simulation, $C=|\mathcal{S}_i|$ and all capacities start at $\%100$. At regular intervals, they are readjusted to support $\alpha C$ contacts (with the headroom set to $\alpha=2$ by default) if there is sufficient change in $C$. Anytime an overflow is detected, this again triggers the same capacity readjustment. Naturally, every such re-adjustment triggers a host-device synchronization, but this auto-calibration tends to quickly converge to a stable state.

\subsubsection{Oriented Bounding-Box Prepass} Before culling spheres within each pair $(i,j)$, an oriented bounding-box (OBB) prepass rejects entire collision pairs.

\section{Methods: Contact Dynamics} \label{methods_cd}
With the collision detection routine covered, this section focuses on the proposed contact dynamics model. We start with a compressed summary of how commonly used simulators in robotics such as MuJoCo \cite{todorov2012MuJoCo} and Drake \cite{drake} operate.

\subsection{Background} The dynamics of a rigid-body system with generalized position $\mathbf q \in \mathbb R^{n_q}$, generalized velocity $\mathbf v \in \mathbb R^{n_v}$, and generalized acceleration $\mathbf a \in \mathbb R^{n_v}$, under a set of $C$ separate external forces $\mathbf{f} \in \mathbb R^{3C}$ (concatenated into a single vector) is governed by the Newton-Euler equation:
\begin{equation}
    \mathbf M(\mathbf q) \mathbf a + \mathbf b(\mathbf q, \mathbf v) = \bm \tau_u + {{\mathbf J(\mathbf q)}}^\trsp \mathbf f
\end{equation}
where $\mathbf M(\mathbf q) \in \mathbb{R}^{n_v \times n_v}$ is the generalized inertia matrix, $\mathbf b(\mathbf q, \mathbf v)$ is the generalized bias force, $\bm \tau_u$ is the generalized force due to actuators, and $\mathbf J(\mathbf q) \in \mathbb{R}^{n_v \times 3C}$ is the concatenation of the Jacobians $\mathbf J_i \in \mathbb R^{3 \times n_v}$ computed at the application points of each individual force $\bm f_i \in \mathbb R^3$ within the concatenated force vector $\mathbf f \in \mathbb R^{3C}$. The expressions $\mathbf a_f = \mathbf M^{-1}(\bm \tau_u - \mathbf b)$ and $\mathbf{v}_f = \mathbf v + \mathbf a_f \Delta t$ denote the free acceleration (i.e. the generalized acceleration in the absence of any external forces) and free velocity (i.e. the generalized velocity after Euler-integration with timestep $\Delta t$ and acceleration $\mathbf a_f$). The generalized acceleration in the presence of external forces can in turn be computed as $\mathbf a = \mathbf{M}^{-1}(\bm \tau_u - \mathbf b) + \mathbf{M}^{-1}{\bar{\mathbf J}}^\trsp \mathbf f = \mathbf a_f + \mathbf{M}^{-1}{\bar{\mathbf J}}^\trsp \mathbf f$. The post-impact generalized velocity $\mathbf v^+$ is then defined via Euler integration as $\mathbf v^+ = \mathbf v + \mathbf a \Delta t = \mathbf v_f + (\mathbf{M}^{-1}{\bar{\mathbf J}}^\trsp) \mathbf f \Delta t$. This means concatenated post-impact contact velocities $\bar{\bm v}^+ \in \mathbb{R}^{3N}$ are in turn defined as $\bar{\bm v}^+ = \bar{\mathbf J}\mathbf v^+ = \bar{\bm v}_f + (\bar{\mathbf J}\mathbf{M}^{-1}{\bar{\mathbf J}}^\trsp) \mathbf f \Delta t = \bar{\bm v}_f + \mathbf G \mathbf f\Delta t$, where $\mathbf G =  \bar{\mathbf J}\mathbf{M}^{-1}{\bar{\mathbf J}}^\trsp \in \mathbb{R}^{3N \times 3N}$ is the Delassus matrix.

MuJoCo \cite{todorov2012MuJoCo} and Drake \cite{castro_sim} solve a second-order cone program (SOCP) to obtain contact forces:
\begin{equation}
    \min_{\mathbf{f}} \: \frac{1}{2}\mathbf{f}^\trsp \left[(\mathbf G  + \mathbf R) \Delta t\right]\mathbf{f} + (\bar{\bm v}_f + \bm \gamma)^\trsp \mathbf f \label{eq:SOCP} \quad \text{s.t.} \quad \mathbf{f} \in \mathbb{K}_{\mu}
\end{equation}
where the regularization matrix $\mathbf R(\mathbf q, \mathbf v) \in \mathbb R^{3C \times 3C}$ ensures that the objective is strictly positive definite, and the term $\bm \gamma(\mathbf q, \mathbf v) \in \mathbb R^{3C}$ implements penetration resolution. When $\mathbf R = \mathbf 0$ and $\bm \gamma = \bm 0$, the optimization problem reduces to finding contact forces within the friction cone that minimize post-impact kinetic energy of the rigid-body system after a semi-implicit Euler update (without resolving any existing penetrations, as this would require injecting energy into the rigid-body system). The proposed contact dynamics formulation described in the upcoming sections can work with arbitrary ways of setting $\mathbf R$ and $\mathbf \gamma$, and the software implementation provides the MuJoCo \cite{todorov2012MuJoCo} and kinetic energy minimization (i.e. $\mathbf R =\mathbf 0$, $\bm \gamma = \bm 0$) conventions by default.

\subsection{Spring-Damper Based Contact Models}
Unlike the optimization-based time-stepping methods of robotics and graphics, simulators commonly used in mechanical engineering \cite{mscadams} express contact forces in closed form as smooth functions of state and integrate the ODE adaptively \cite{kurtz2025cenic}.
These closed form expressions often take a functional form analogous to a spring-damper with dissipation (hence our naming convention for this contact model class), and are well-validated through empirical measurements in the real world \cite{kurtz2025cenic, johnson1987contact}.

The main downside of the spring-damper based approach is that time-adaptive integration does not readily lend itself to massive vectorization, and rollouts have to progress and terminate at the speed of the stiffest simulation instance in the batch \cite{diffrax}. And without time-adaptive integration, creating a stable simulation requires careful hand-tuning for every new scene and interaction modality (e.g. anticipating the motions that will be involved, states that may be visited, and the resulting inertias that will be reflected on each contact point). On the other hand, with discrete time-stepping methods (to quote Todorov et al. \cite{todorov2014convex} verbatim) \say{one can think of the contact impulses as being generated by smart spring-dampers, which are aware of each other and furthermore scale their stiffness and damping automatically with inertia}. As explained in the next section, this intuition forms the basis of the contact models proposed in this paper, which operate by computing spring-damper based contact forces between collision pairs and then rescaling them (with a decision variable per collision pair) in a way that optimizes the original SOCP objective \eqref{eq:SOCP}. In particular, the proposed framework provides two spring-damper based contact model options for this purpose: i) the smoothed linear spring model with Hunt and Crossley dissipation as proposed in \cite{kurtz2023inverse}, and ii) the \say{complementarity-free} spring-damper model of \cite{borse2026comfree} with inertia-aware stiffness and damping terms.

\subsection{Proposed Contact Model}\label{sec:contact-model}
Let us continue our exposition with the mathematical details of the proposed contact model.
\subsubsection{Rescaling Spring-Damper Forces} Let us denote the set of all collision pairs as $\mathcal{K}$, indexed with $k \in \{1, \cdots, |\mathcal K|\}$. Let us further denote the total number of contact points in each collision pair as $C_k$.
Finally, let us use the notation $\bm f_{k, l}(\mathbf q, \mathbf v): \mathbb R^{n_q + n_v} \to \mathbb R^3$ to denote the spring-damper contact force computed for the $l$'th contact point of the collision pair $k$, as a function of system position and velocity. As previously defined, $\mathbf{f} \in \mathbb R^{3C}$ denotes the concatenation of individual contact forces $\bm f_{k, l} \in \mathbb R^3$, with the total number of contacts satisfying $C = \sum_{k=1}^{|\mathcal K|} C_k$. It is then possible to transcribe a lumped parameterization of $\mathbf{f}$ in terms of per-collision-pair scaling factors $\bm \alpha \in \mathbb R^{|\mathcal{K}|}$ as $\mathbf{f}(\bm \alpha) = \mathbf F \bm \alpha$, where $\mathbf F(\mathbf q, \mathbf v) \in \mathbb R^{3C \times |\mathcal K|}$ simply consists of the vectors $\bm f_{k, l}(\mathbf q, \mathbf v)$ rearranged in a matrix form so that they share a single scaling factor $\alpha_k$ within the product $\mathbf F \bm \alpha$. The key observation is that under this lumped parameterization, the SOCP \eqref{eq:SOCP} becomes:
\begin{align} \label{eq:bqp}
    \min_{\bm \alpha} &\: \frac{1}{2}\mathbf{f}(\bm \alpha)^\trsp \left[(\mathbf G  + \mathbf R) \Delta t\right]\mathbf{f}(\bm \alpha) + (\bar{\bm v}_f + \bm \gamma)^\trsp \mathbf f(\bm \alpha) \\
    \text{s.t.}  &\: \bm{\alpha} \succeq \bm 0 \nonumber
\end{align}
which is a box-constrained QP (BQP) since $\mathbf{f}(\bm \alpha) = \mathbf F \bm \alpha$ is affine. In particular, as spring-damper forces lie within the friction cone by closed form construction, the cone constraints $\mathbf{f} \in \mathbb{K}_{\mu}$ simply become a box constraint $\bm{\alpha} \succeq \bm 0$.

In the software implementation, the BQP \eqref{eq:bqp} is solved via projected coordinate descent, with each parallel simulation instance running the algorithm in a single thread (i.e. without any intra-simulation parallelism or any cross-thread synchronization). There are two main reasons for this. First, it allows parallelization of QPs with different numbers of decision variables (as each simulation in the batch can have a different number of collision pairs active that survive the OBB prepass). And second, it can handle the case where $\bm Q$ is only positive semidefinite, since this causes a zero diagonal entry $Q_{ii} = 0$, which in turn means the coordinate descent update for the single coordinate $i$ actually sees a linear objective and is hence reduced to a box projection depending on $\text{sign}(c_i)$.

\subsubsection{Stagger Iterations} After computing the scaling factors $\bm \alpha$, the generalized acceleration under their influence is simply $\mathbf a = \mathbf a_f + \mathbf{M}^{-1}{\bar{\mathbf J}}^\trsp \mathbf F \bm \alpha$, meaning their contribution to the generalized acceleration can be isolated as $\Delta \mathbf a(\bm \alpha) = \mathbf{M}^{-1}{\bar{\mathbf J}}^\trsp \mathbf F \bm \alpha$. With this observation in place, we can extend the spring-damper rescaling formulation from the previous section into the following stagger scheme:
\begin{enumerate}[leftmargin=*, label={\color{ourblue}(\roman*)}]
    \item Initialize $\mathbf{f}_{\text{tot}} = \bm 0$, $\mathbf v^+ = \mathbf{v}_f$, $\mathbf F = \mathbf F(\mathbf q, \mathbf v^+)$, and define the parameterization $\mathbf{f}(\bm \alpha) = \mathbf F \bm \alpha + \mathbf{f}_{\text{tot}}$.
    \item Solve BQP \eqref{eq:bqp} with $\mathbf{f}(\bm \alpha) = \mathbf F \bm \alpha + \mathbf{f}_{\text{tot}}$ to obtain $\bm \alpha^*$.
    \item Update $\mathbf{f}_{\text{tot}} = \mathbf{f}_{\text{tot}} + \mathbf F \bm \alpha^*$ and $\mathbf v^+ = \mathbf v^+ + \Delta \mathbf a(\bm \alpha^*)\Delta t$.
    \item Obtain a new spring-damper force distribution $\mathbf F = \mathbf F(\mathbf q, \mathbf v^+)$ under the new velocity $\mathbf v^+$.
    \item Iterate (ii)-(iv) until convergence.
\end{enumerate}
In the software implementation, stagger iterations are run using the norm of $\Delta \mathbf a(\bm \alpha^*)$ as a convergence criterion.

The intuition behind this staggering scheme is as follows. The original BQP formulation \eqref{eq:bqp} assigns only a single scaling factor per collision pair, even though every collision pair under the proposed collision detection routine produces a dense manifold of multiple contact points. Not having a per-contact point decision variable inherently imposes a restriction on how expressive (i.e. multimodal) the contact force distribution between a collision pair can be, primarily determined by the functional form of the spring-damper model $\bm f_{k, l}(\mathbf q, \mathbf v)$. The staggering scheme addresses this limitation by building the total contact force $\mathbf{f}_{\text{tot}}$ additively across multiple iterations, where the spring-damper force distribution is allowed to readjust at each iteration. And this re-adjustment is based on the contact velocities $\mathbf v^+$ that would be observed if one were to do a semi-implicit Euler update with the contact force $\mathbf{f}_{\text{tot}}$ accumulated so far.
For example, if the accumulated contact force only pushes out one side of a contact manifold, the closed-form spring-damper expression in the next iteration assigns that side no weight (under the time-stepped velocity $\mathbf v^+$ seen by the current stagger iteration, it \emph{is} pushed out). The distribution then concentrates on the remaining side, introducing a new mode.

\subsubsection{Frictional Correction} As previously described, adding a regularization matrix $\mathbf R \neq \bm 0$ inadvertently introduces a $\mathbf f^\trsp \mathbf R \mathbf f$ term into the optimization objective that scales down contact forces. This means frictional contact forces cannot achieve the exact value needed to maintain complete staticity in a rigid-body system (e.g. zero velocity and acceleration), meaning all contacts eventually slip under constant forces such as gravity. This artifact is present in most commonly used simulators in robotics \cite{todorov2012MuJoCo, drake, tsounis2026kamino}. The proposed contact dynamics model fixes it via the following additions to the stagger iterations:
\begin{itemize}
    \item A tangential direction $\hat{\bm t}_i$ is constructed for each contact point $i$. The software implementation provides two choices, normalizing the accumulated: i) tangential contact force $\mathbf f_{\text{tot}, i, t}$, or ii)  tangential contact velocity $\bm J_i \mathbf v_t^+ \in \mathbb R^3$.
    \item For each contact point $i$, the gap between $\mathbf{f}_{\text{tot}, i, t}$ and its Coulomb upper bound $\mu_i \mathbf{f}_{\text{tot}, i, n}$ (using the current accumulated normal force) is computed as $\Delta f_{i, t} = \mu_i \mathbf{f}_{\text{tot}, i, n} - \mathbf{f}_{\text{tot}, i, t}$.
    \item The frictional correction vectors $\Delta f_{i, t} \hat{\bm t}_i$ are constructed for each contact point, and are re-arranged in a matrix form $\bar {\mathbf F}$ so that they share a single scaling factor $\alpha_k$ (per collision pair $k$) within the product $\bar {\mathbf F} \bm \alpha$ (same as spring-damper forces from the previous section).
\end{itemize}

With this construction in place, let us define the affine parameterization $\bar{\mathbf f}(\bm \alpha) = \mathbf{f}_{\text{tot}} + \bar{\mathbf F} \bm \alpha$ to use for frictional correction. Note that setting all scaling factors to one results in a contact force $\bar{\mathbf f}(\bm 1)$ where all points saturate their Coulomb upper-bound (exactly if $\hat{\bm t}_i$ comes from $\mathbf{f}_{\text{tot}, i, t}$, only approximately if it comes from $\bm J_i \mathbf v^+ $). One can then compute the frictional correction that minimizes the system kinetic energy after a semi-implicit Euler update by solving the following BQP:
\begin{align}
    \min_{\bm \alpha} \:& \frac{1}{2}\bar{\mathbf f}(\bm \alpha)^\trsp (\mathbf G  \Delta t)\bar{\mathbf f}(\bm \alpha) + \bar{\bm v}_f^\trsp \ \bar{\mathbf f}(\bm \alpha) \label{eq:bqp_kinmin} \\
    \text{s.t.} \: & \bm 0 \preceq \bm{\alpha} \preceq \bm 1 \nonumber
\end{align}

Overall, this yields the following stagger scheme:
\begin{enumerate}[leftmargin=*, label={\color{ourblue}(\roman*)}]
    \item Initialize $\mathbf{f}_{\text{tot}} = \bm 0$, $\mathbf v^+ = \mathbf{v}_f$, $\mathbf F = \mathbf F(\mathbf q, \mathbf v^+)$, $\bar {\mathbf F} = \bm 0$.
    \item Solve BQP \eqref{eq:bqp} with $\mathbf{f}(\bm \alpha) = \mathbf F \bm \alpha + \mathbf{f}_{\text{tot}}$ to obtain $\bm \alpha^*$.
    \item Update $\mathbf{f}_{\text{tot}} = \mathbf{f}_{\text{tot}} + \mathbf F \bm \alpha^*$ and $\mathbf v^+ = \mathbf v^+ + \Delta \mathbf a(\bm \alpha^*)\Delta t$.
    \item Construct $\bar {\mathbf F}$ for frictional correction, using $\mathbf{f}_{\text{tot}}$ or $\mathbf v^+$.
    \item Solve BQP \eqref{eq:bqp_kinmin} with $\mathbf{f}(\bm \alpha) = \bar {\mathbf F} \bm \alpha + \mathbf{f}_{\text{tot}}$ to obtain $\bm \alpha^*$.
    \item Update $\mathbf{f}_{\text{tot}} = \mathbf{f}_{\text{tot}} + \bar {\mathbf F} \bm \alpha^*$ and $\mathbf v^+ = \mathbf v^+ + \Delta \mathbf a(\bm \alpha^*)\Delta t$.
    \item Obtain a new spring-damper force distribution $\mathbf F = \mathbf F(\mathbf q, \mathbf v^+)$ under the new velocity $\mathbf v^+$.
    \item Iterate (ii)-(vii) until convergence.
\end{enumerate}

An important observation is that the updates in steps (iii) and (vi) allow the spring-damper re-distribution and re-scaling mechanisms and the frictional correction mechanism to converge in a way that is aware of one another.

\subsubsection{Validity of the Lumped Parameterization} Whenever $\mathbf{R} \neq 0$, the SOCP \eqref{eq:SOCP} is exactly equivalent to the following unconstrained non-linear least-squares problem on the generalized acceleration $\mathbf a$ (often called its \say{primal form} \cite{todorov2012MuJoCo, castro_sim}):
\begin{align}
    \min_{\mathbf{a}} \:& \frac{1}{2}(\mathbf{a} - \mathbf{a}_f)^\trsp \mathbf M (\mathbf{a} - \mathbf{a}_f) \nonumber\\ + &\frac{1}{2} \left[\mathbf{P}_{\mathbb K_\mu}(\bar{\bm a}_\text{ref} - \bar{\mathbf{J}}\mathbf a)\right]^\trsp \mathbf R^{-1}  \left[\mathbf{P}_{\mathbb K_\mu}(\bar{\bm a}_\text{ref} - \bar{\mathbf{J}}\mathbf a)\right] \label{eq:primal_generic}
\end{align}
where $\mathbf{P}_{\mathbb K_\mu}$ is the closed form projection operator onto the 3D circular cone, and $\bar{\bm a}_\text{ref} = \bar{\mathbf{J}}\mathbf a_f - {\mathbf r}/{\Delta t} \in \mathbb{R}^3$. This means even if the apparent dimensionality of the SOCP \eqref{eq:SOCP} is $3C$, the intrinsic mechanism simply operates by adjusting the $n_v$ degrees of freedom between all colliding bodies such that a least-squares fit is achieved with respect to contact acceleration references $\bar{\bm a}_\text{ref}$ obtained from closed form spring-damper expressions. This means the SOCP formulation \eqref{eq:SOCP} also does not make decoupled adjustments between contact points that lie on the same body and are in close proximity (since they share the same underlying degrees of freedom within $\mathbf a$). Therefore, the main difference between the well-established SOCP formulation and the lumped parametrization presented here is how they choose to introduce a coupling between proximate contact points: the former couples contact points that share the same degrees of freedom in the multi-body system, and the latter couples contact points that share the same collision pair.

\section{Experiments} \label{experiments}
Let us now present a set of experiments that characterize the proposed simulator across three axes: i) physical faithfulness, ii) throughput, and iii) usefulness for contact-rich control. Throughout all experiments, {\color{ourblue} Lodos-MJ-N} denotes the pipeline proposed in this paper, optimizing the MuJoCo convention for $\mathbf R$ and $\mathbf \gamma$, with N stagger iterations ($N=0$ reduces to combining the unscaled spring-damper forces obtained by the method of \cite{jin2024complementarity, borse2026comfree} with our collision detection).  {\color{ourblue}MuJoCo-CPU} and {\color{ourblue}MJWarp} denote the default CPU implementation and the GPU implementation \cite{MuJoCo_warp} in NVIDIA Warp \cite{Macklin_Warp_A_High-performance_2022}; {\color{ourblue}MuJoCo-LC} combines our collision detection with MuJoCo contact dynamics. For all experiments and simulators, we use a matched $2\,\mathrm{ms}$ timestep size which is the MuJoCo default value. Simulator settings, including Lodos memory capacities and MJWarp contact and constraint limits, are tuned for maximum performance in each experiment. All experiments run on a single NVIDIA RTX 4080 GPU. Supplementary videos accompany all experiments.

\begin{figure} %
    \centering
    \includegraphics[width=\linewidth]{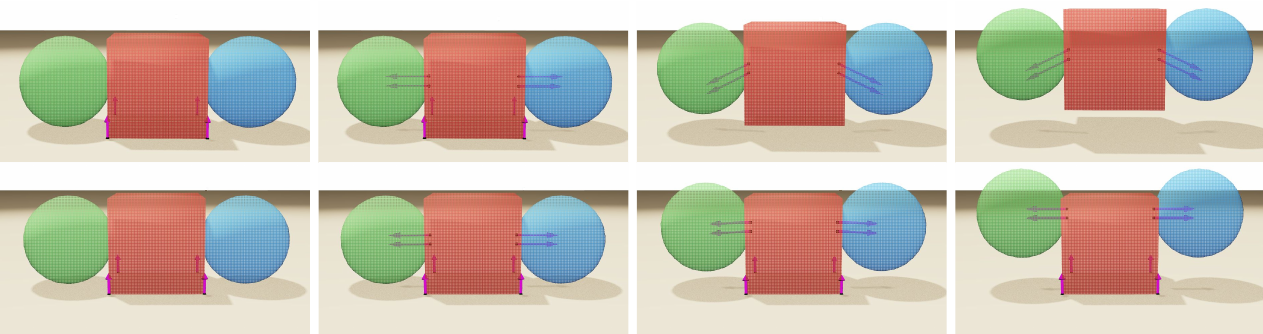}
    \caption{Antipodal pinching and lifting of a cube, with (top) and without (bottom) the proposed friction correction scheme.
    }
    \label{fig:pinch}%
\end{figure}

\begin{figure}
    \centering
    \includegraphics[width=\linewidth]{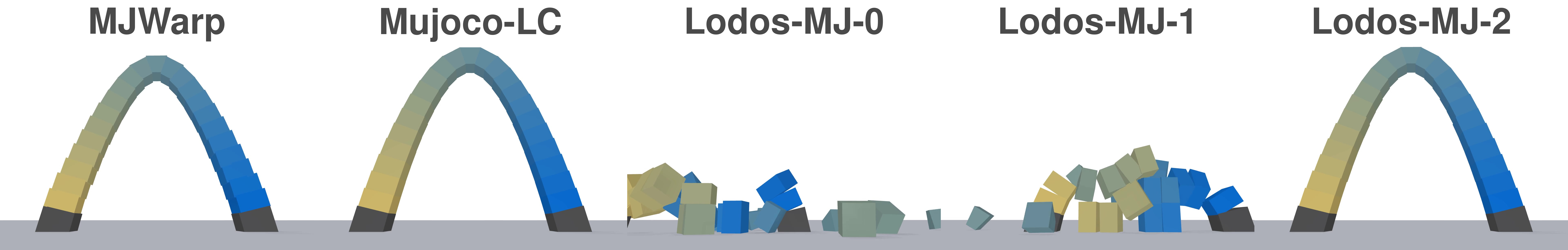}
    \caption{Final frames from 15-second simulations of a masonry arch.
    }
    \vspace{-2mm}
    \label{fig:arch}
\end{figure}

\subsection{Physical Faithfulness}

\subsubsection{Effectiveness of the Friction Correction Scheme} Fig. \ref{fig:pinch} conducts a didactic experiment where two spherical fingers with $r=0.5m$ are pinching a $1m$ $10kg$ cube, lifting it up $0.3\,\mathrm m$ over a $1.2\,\mathrm s$ interval with a speed of $0.25\,\mathrm{m/s}$, and holding it for $5s$. The spheres use motor actuators along the horizontal axis applying a $100N$ force each, and position controllers with $k_p=1e^4$ and $k_v=200$ in the vertical axis. The friction coefficient is $\mu=0.6$, meaning the Coulomb upper-bound for the tangential force is $120N$ compared to the $100N$ weight of the cube. The experiment compares the closed form spring-damper model of \cite{jin2024complementarity, borse2026comfree} with and without our frictional correction scheme. Without frictional correction, the deliberately small rising velocity of the fingers prevents the spring-damper model from generating sufficient frictional force to lift the cube (the maximum observed value is $8.22\,\mathrm N$). With our friction correction scheme, the fingers exert an exact $120N$ frictional force on the cube during the first $114 ms$ of the pickup resulting in $5.97\,\mathrm mm$ of slip in between, after which point the contact sticks. During the $5s$ hold afterwards, a mean creep velocity of $-0.0014 mm/s$ is observed under $g=9.81\,m/s^2$ gravitational acceleration (likely due to the finite tolerance and 32-bit precision of our BQP solver). These results show that the proposed frictional correction scheme operates as intended.
\subsubsection{Stability of a Masonry Arch} Fig. \ref{fig:arch} shows a masonry arch simulation with $25$ stones. As can be seen, unscaled spring-damper forces alone cannot achieve sufficient physical fidelity to hold the arch upright. Doing one stagger iteration to rescale the spring-damper forces diminishes the severity, and by two stagger iterations the arch is stable. Stability of the MuJoCo-LC baseline further confirms the validity of the proposed collision detection method. In particular, MJWarp's reliance on GJK+EPA contact points that jump discontinuously across timesteps (due to parallel-face collisions) results in visible slipping between the stones, an artifact fixed by the MuJoCo-LC and Lodos-MJ-2 configurations.

\subsubsection{Locomotion with a Bipedal Robot}
Fig. \ref{fig:overview} shows successful sim-to-sim transfer for the “Microduck” platform, where policies trained in MJWarp (copied directly from the open-source Microduck repository \cite{pollen_robotics_microduck_rl}) are deployed zero-shot in Lodos to execute diverse locomotion behaviors. This further demonstrates the physical faithfulness of Lodos.\footnote{Please refer to the supplementary for videos of all simulations.}

\begin{figure}
    \centering
    \includegraphics[width=\linewidth]{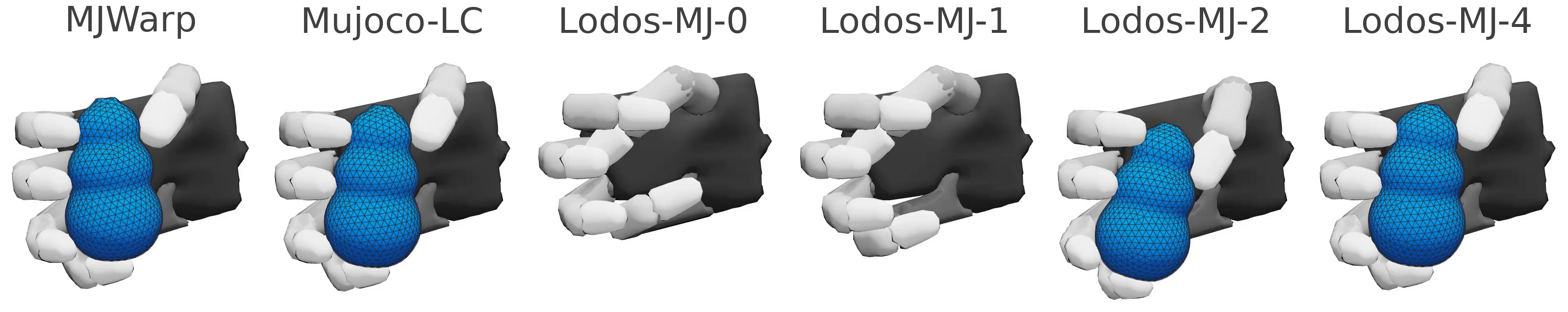}\\
    \caption{
    Final frames from a $5\,\mathrm{s}$ simulation in which a position controller, with parameters optimized in MuJoCo to maintain a delicate prismatic grasp, is transferred directly to Lodos.
    }
    \vspace{-4mm}
    \label{fig:dexonomy}
\end{figure}

\subsubsection{Prismatic Grasp with an Anthropomorphic Hand} Fig.\ref{fig:dexonomy} shows another successful sim-to-sim transfer, this time of a delicate four-finger prismatic grasp. The grasping pose and the position control targets that squeeze it are optimized in MuJoCo using the Dexonomy framework \cite{chen2025dexonomy}, and are then directly transferred to Lodos. The results show the same pattern as the arch experiment, where physical faithfulness increases with further stagger iterations.

\subsection{Throughput Characterization}
\input{throughput_table_fused}
\begin{figure}
    \centering
    \includegraphics[width=0.99\linewidth]{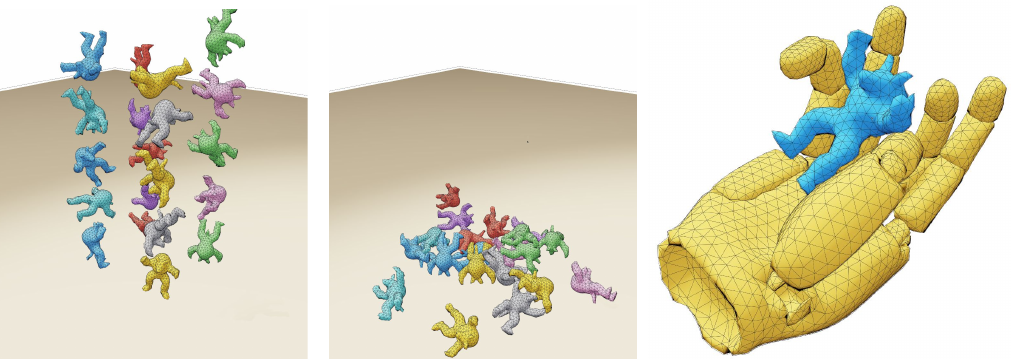}\\
    \caption{Throughput characterization experiments that involve complex mesh-mesh collisions. \textbf{Left}: $20$ armadillos falling on a plane. \textbf{Right}: A Sharpa Wave hand holding an armadillo.
    }
    \label{fig:amrdillo-sharpa-throughput}
    \vspace{-4mm}
\end{figure}

\subsubsection{Simulating Complex Collision Geometry} Figure~\ref{fig:amrdillo-sharpa-throughput} and Table~\ref{tab:throughput} evaluate batches of 20 armadillo meshes falling onto a plane, with randomized initial configurations per instance. Lodos-MJ-1 represents each armadillo with $451$ vertices and $100$ spheres; MJWarp uses $34$ convex polytopes. Throughput is averaged over the $1$--$4\,\mathrm s$ interval of a $10\,\mathrm s$ simulation, when impacts and contacts dominate. This excludes contact-free and settled phases where solver warm starts confound comparison. Lodos maintains physical faithfulness while achieving an order of magnitude higher peak throughput.
\subsubsection{In-Hand Manipulation with Detailed Meshes}\label{sec:inhand_bench} We also benchmark a Sharpa Wave hand manipulating the same armadillo (Fig.~\ref{fig:amrdillo-sharpa-throughput}, Table~\ref{tab:throughput}). Lodos-MJ-1 represents the hand with $3{,}479$ vertices across $23$ bodies, without any inscribed spheres (as placing them only within the armadillo is sufficient). MJWarp in turn uses 87 convex polytopes for the entire hand. To evaluate throughput across varying batch sizes, we run the MPPI controller from S.\ref{sec:mppi}, tasked with tracking four random goals sampled to be $60^\circ$ apart along a random rotation vector, to generate a full contact-rich manipulation trajectory. We then extract the $2,500$ states with the highest number of contacts. Each batch with size $B$ across the ladder is constructed by sampling $B$ states out of these $2500$ with replacement. The reported values are averaged over three repetitions. The results demonstrate that the proposed framework again achieves an order of magnitude higher maximum throughput compared to MJWarp.

\subsection{Usefulness for Contact-Rich Control}
\begin{figure}
    \centering
    \includegraphics[width=0.99\linewidth]{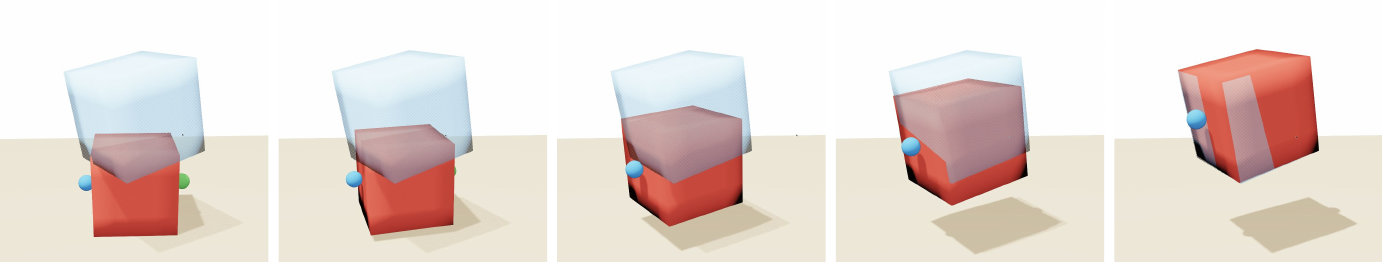}\\
    \caption{Trajectory optimization experiment using iLQR, with linearizations of the system dynamics estimated via zeroth-order randomized smoothing using Lodos-MJ-1.
    }
    \label{fig:ilqr}
    \vspace{-5mm}
\end{figure}
\subsubsection{iLQR with Randomized Smoothing} Fig.\ref{fig:ilqr} shows a trajectory optimization example using iLQR, where linearizations use the zeroth-order randomized smoothing method of \cite{pang2023global}. The analogous experiment in \cite{pang2023global} pushes a cube with a sphere on a 2D plane (i.e. each with two DoF) under quasistatic dynamics without gravity or ground friction. Here, two position-controlled three-DoF spheres ($r=0.1\,\mathrm m$, $1\,\mathrm{kg}$) pinch and lift a six-DoF cube ($1\,\mathrm m$, $5\,\mathrm{kg}$) by $1\,\mathrm m$ with a random $60^\circ$ target rotation, under full second-order dynamics and gravity. The state space (generalized position and velocity) is $24$ dimensional and the rotational degree of freedom uses the so(3) parameterization. Controls are $6$ dimensional, and the parameterization splits the $2s$ trajectory into $20$ knot points interpolated with a zero-order hold. The objective and $Q,R$ matrices follow \cite{pang2023global}. Noise level for randomized smoothing is picked via grid search on the iLQR termination cost.

The experiment involves two configurations. In the first one, both the dynamics linearization and the iLQR rollouts (and line-search) use Lodos-MJ-1. The second configuration is mixed, where linearization uses Lodos-MJ-1 and rollouts use MJWarp. Both successfully reach the goal configuration with very similar precision, with the former achieving a $0.75mm$ translational and $0.13^\circ$ rotational error, and the latter $10.5\,\mathrm{mm}$ and	$1.24^\circ$ respectively. This demonstrates that the proposed contact model can be used to estimate gradients that are useful for trajectory optimization, which remain informative even when transferred across different contact models.

\begin{figure}
    \centering
    \includegraphics[width=0.99\linewidth]{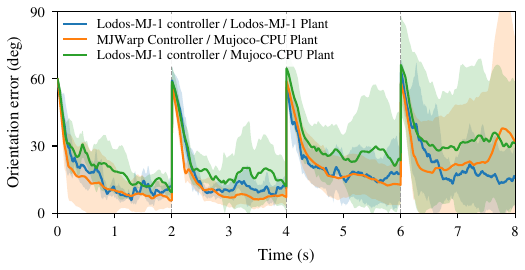}\\
    \vspace{-2mm}
    \caption{Stochastic MPC experiment where MPPI is used to manipulate an armadillo in hand with a Sharpa Wave hand, tracking four consecutive goal orientations. The plots show the mean and standard deviation across $5$ randomly sampled goal sequences.
    }
    \label{fig:mppi_plot}
    \vspace{-5mm}
\end{figure}

\subsubsection{In-Hand Manipulation with MPPI}\label{sec:mppi} Fig. \ref{fig:mppi_plot} shows the results of an in-hand manipulation experiment with the same Sharpa Wave hand and armadillo from S.\ref{sec:inhand_bench}. The SO(3) pose of the armadillo needs to track four consecutive goal poses that differ by a $60^\circ$ rotation along a random axis in 3D. Each goal is held fixed for a quarter of the overall $8\,\mathrm{s}$ manipulation sequence. All experiments use an MPPI controller with 64 parallel rollouts and without any domain randomization, as implemented by the hydrax framework \cite{kurtz2024hydrax} (where all the remaining MPPI parameters are directly carried over from the analogous in-hand manipulation experiment in hydrax). There are three simulator configurations which are used for the plant and for obtaining the rollouts of the MPPI controller: i) Lodos-MJ-1 for both the controller and the plant, ii) MJWarp for the controller and MuJoCo-CPU for the plant, and iii) Lodos-MJ-1 for the controller and MuJoCo-CPU for the plant. The tracking performance plots and the supplementary videos demonstrate that: i) Lodos-MJ-1 can simulate extremely fast, contact-rich manipulation behavior without artifacts, and ii) actions optimized in Lodos-MJ-1 successfully transfer to MuJoCo-CPU.

\section{Limitations and Future Work} \label{limitations}
An important clarification about the proposed framework is that it is designed to maximize \emph{throughput} in the presence of \emph{complex mesh-mesh collisions}. For small (e.g. less than $\sim 256$) or medium-sized (e.g. $\sim 256 - 1024$) batches, the rollout APIs of MuJoCo-CPU and MJWarp remain the best options respectively. Only by allowing the batch size to be increased does Lodos gain an edge in throughput, and mainly thanks to its collision detection and contact solver routines being designed specifically such that they scale well into this regime. Similarly, when it is sufficient to approximate geometries by primitives (e.g. spheres, capsules, boxes), MJWarp still has the edge in terms of throughput. Another limitation is that the software implementation does currently not support automatic differentiation via Nvidia Warp.

\section{Conclusion} \label{conclusion}
We presented a novel simulation approach built from the ground up to facilitate research on contact-rich manipulation and control. We experimentally validated its merits along three axes: i) physical faithfulness, ii) computational throughput, and iii) utility for control. Promising directions for future work include extending the framework to support automatic differentiation and leveraging it to develop novel optimal control algorithms for contact-rich settings.

\section*{Acknowledgments}
Software development was assisted by Claude Opus 5.

\bibliographystyle{IEEEtranN} %
\footnotesize
\bibliography{lodos}

\end{document}

%% file: header.tex
\usepackage[utf8]{inputenc} %
\usepackage[T1]{fontenc}    %
\usepackage{mathptmx}       %
\usepackage{times}
\usepackage{microtype}      %

\usepackage{graphicx}
\usepackage{epsfig}         %
\usepackage{adjustbox}      %
\usepackage{tikz}

\usepackage{booktabs}       %
\usepackage{tabularx}
\usepackage{threeparttable}
\usepackage{multicol}

\usepackage{amsmath}
\usepackage{amsfonts}       %
\usepackage{amssymb}
\usepackage{bbm}
\usepackage{bm}
\usepackage{nicefrac}       %
\usepackage[makeroom]{cancel}

\usepackage{url}            %
\usepackage{xspace}
\usepackage{dirtytalk}
\usepackage{soul}
\usepackage{epigraph}
\usepackage{comment}
\usepackage[font=small,labelfont=bf]{caption}

\makeatletter
\let\labelindent\undefined
\makeatother
\usepackage[shortlabels]{enumitem}

\makeatletter
\let\NAT@parse\undefined
\makeatother
\usepackage[numbers]{natbib}

\usepackage{xcolor}

\definecolor{ourblue}{RGB}{91,135,201} %
\definecolor{ourorange}{rgb}{0.881,0.611,0.142} %
\definecolor{ourgreen}{rgb}{0.56,0.692,0.195}   %
\definecolor{ourred}{rgb}{0.923,0.386,0.209}    %
\definecolor{ourviolet}{rgb}{0.528,0.471,0.701} %
\definecolor{ourbrown}{rgb}{0.772,0.432,0.102}  %
\definecolor{ourazure}{rgb}{0.364,0.619,0.782}  %
\definecolor{ourolive}{rgb}{0.572,0.586,0.}     %
\definecolor{ourgray}{RGB}{102,88,84}           %

\definecolor{ourblue2}{RGB}{9,134,223} %
\definecolor{ourdarkblue2}{RGB}{5,97,164} %
\definecolor{ourlightblue2}{RGB}{132,201,250} %

\definecolor{ourorange2}{RGB}{224,90,18} %
\definecolor{ourdarkorange2}{RGB}{160,63,9} %
\definecolor{ourlightorange2}{RGB}{246,175,137} %

\definecolor{ouryellow2}{RGB}{227,213,25} %
\definecolor{ourdarkyellow2}{RGB}{177,166,17} %
\definecolor{ourlightyellow2}{RGB}{242,235,140} %

\definecolor{ourpink2}{RGB}{247,24,139} %
\definecolor{ourdarkpink2}{RGB}{164,4,86} %
\definecolor{ourlightpink2}{RGB}{250,163,207} %

\definecolor{ourgreen2}{RGB}{159,198,52} %
\definecolor{ourdarkgreen2}{RGB}{109,138,30} %
\definecolor{ourlightgreen2}{RGB}{209,228,154} %

\definecolor{ourgray2}{RGB}{124,124,115} %
\definecolor{ourdarkgray2}{RGB}{87,87,81} %
\definecolor{ourlightgray2}{RGB}{194,194,189} %

\newcommand{\ouritemizelabel}{\protect\textcolor{ourblue}{\textbf{\textbullet}}}
\setlist[itemize]{leftmargin=*, label={\ouritemizelabel}}

\makeatletter
\DeclareRobustCommand\onedot{\futurelet\@let@token\@onedot}
\def\@onedot{\ifx\@let@token.\else.\null\fi\xspace}
\makeatother

\newcommand{\trsp}{{\scriptscriptstyle\top}}

\makeatletter
\newcommand*{\addFileDependency}[1]{%
  \typeout{(#1)}
  \@addtofilelist{#1}
  \IfFileExists{#1}{}{\typeout{No file #1.}}
}
\makeatother

\usepackage{xr-hyper}
\usepackage[breaklinks,colorlinks,allcolors=ourblue]{hyperref}
\usepackage{cleveref}

%% file: throughput_table_fused.tex
\providecommand{\contax}{\textsc{contax}}
\providecommand{\mjsd}{\textsc{Lodos-MJ-N}}
\providecommand{\mjwarp}{MJWarp}
\providecommand{\nstag}{N_\mathrm{s}}
\providecommand{\textmu}{\ensuremath{\mu}}
\begin{table*}[t]
  \centering
  \begin{threeparttable}
  \caption{Simulation throughput across batch sizes, and speed-up over \mjwarp{}.}
  \label{tab:throughput}
  \small
  \setlength{\tabcolsep}{3.5pt}
  \begin{tabular}{llc*{9}{r@{}l}@{\hspace{1.5em}}cc}
    \toprule
    & & & \multicolumn{18}{c}{Step time per world (\textmu{}s) at batch size $B$} & \multicolumn{2}{c}{Speed-up over \mjwarp{}} \\
    \cmidrule(lr){4-21} \cmidrule(lr){22-23}
    Scene & Simulator & $\nstag$ & \multicolumn{2}{c}{$2^{0}$} & \multicolumn{2}{c}{$2^{3}$} & \multicolumn{2}{c}{$2^{6}$} & \multicolumn{2}{c}{$2^{9}$} & \multicolumn{2}{c}{$2^{12}$} & \multicolumn{2}{c}{$2^{13}$} & \multicolumn{2}{c}{$2^{14}$} & \multicolumn{2}{c}{$2^{15}$} & \multicolumn{2}{c}{$2^{16}$} & \begin{tabular}[b]{@{}c@{}}Whole\\step\end{tabular} & \begin{tabular}[b]{@{}c@{}}Contact\\pipeline\end{tabular} \\
    \midrule
    In-hand & \mjwarp{} & & 904& & 147& & 21&.8 & 4&.08 & \textbf{2}&\textbf{.75} & \multicolumn{2}{c}{---} & \multicolumn{2}{c}{---} & \multicolumn{2}{c}{---} & \multicolumn{2}{c}{---} & & \\
    reorientation & \mjsd{} & 0 & 523& & 74&.3 & 10&.7 & 2&.12 & 0&.438 & 0&.317 & \textbf{0}&\textbf{.299} & 0&.313 & 0&.346 & 9.19$\times$ & 14.0$\times$ \\
    & & 1 & 656& & 102& & 15&.9 & 3&.07 & 0&.590 & 0&.409 & \textbf{0}&\textbf{.367} & 0&.374 & 0&.419 & 7.49$\times$ & 11.0$\times$ \\
    & & 2 & 799& & 136& & 21&.3 & 3&.91 & 0&.733 & 0&.501 & 0&.440 & \textbf{0}&\textbf{.437} & 0&.495 & 6.29$\times$ & 9.90$\times$ \\
    \midrule
    20-armadillo & \mjwarp{} & & 1335& & 216& & 40&.1 & \textbf{17}&\textbf{.9} & \multicolumn{2}{c}{---} & \multicolumn{2}{c}{---} & \multicolumn{2}{c}{---} & \multicolumn{2}{c}{---} & \multicolumn{2}{c}{---} & & \\
    drop & \mjsd{} & 0 & 270& & 54&.4 & 14&.7 & 3&.02 & 0&.942 & \textbf{0}&\textbf{.824} & 0&.879 & \multicolumn{2}{c}{---} & \multicolumn{2}{c}{---} & 21.7$\times$ & 24.8$\times$ \\
    & & 1 & 404& & 84&.1 & 24&.7 & 4&.96 & 1&.33 & \textbf{1}&\textbf{.21} & 1&.40 & \multicolumn{2}{c}{---} & \multicolumn{2}{c}{---} & 14.8$\times$ & 16.7$\times$ \\
    & & 2 & 580& & 134& & 39&.5 & 7&.76 & 1&.96 & \textbf{1}&\textbf{.73} & 2&.00 & \multicolumn{2}{c}{---} & \multicolumn{2}{c}{---} & 10.3$\times$ & 11.1$\times$ \\
    \bottomrule
  \end{tabular}
  \begin{tablenotes}[flushleft]
    \footnotesize
    \item[]\! Numerical results for the throughput characterization experiments. \textbf{Left:} Wall-clock time of one simulation step divided by the
      batch size.
      Each entry is the median over three batches.
      \textbf{Right:} Speed-up over \mjwarp{} with each simulator at
      its peak batch size, for the whole step and for the contact pipeline alone (i.e. the latter omits the shared unconstrained rigid-body dynamics pass). $\nstag$ denotes the number of stagger iterations.
  \end{tablenotes}
  \end{threeparttable}
  \vspace{-5mm}
\end{table*}